\documentclass{article}

\usepackage{arxiv}
\usepackage[utf8]{inputenc} 
\usepackage[T1]{fontenc}    
\usepackage{url}            
\usepackage{booktabs}       
\usepackage{amsfonts}       
\usepackage{nicefrac}       
\usepackage{microtype}      
\usepackage{lipsum}
\usepackage{graphicx}
\usepackage{natbib}
\usepackage{enumitem}
\usepackage{amsmath}
\usepackage{amssymb}
\usepackage{mathtools}
\usepackage{amsthm}
\usepackage{multirow}
\usepackage{algorithm}      
\usepackage{algorithmic}    

\graphicspath{ {./images/} }
\usepackage{xspace} 
\newcommand{\REPO}{REPO\xspace}
\newcommand{\fTBD}[1]{\textcolor{red}{\emph{(#1)}}}
\newcommand{\KL}{\mathrm{KL}}
\newcommand{\softmax}[1]{\sigma\left(#1\right)}
\renewcommand{\fTBD}[1]{}
\usepackage[table,dvipsnames]{xcolor}
\usepackage[bookmarks=false]{hyperref}
\hypersetup{
	linktocpage=true,
	colorlinks=true,				
	linkcolor=Blue,				
	citecolor=Blue,				
	urlcolor=Blue,				
}
\usepackage[capitalise]{cleveref}

\title{Detoxifying LLMs via Representation Erasure-based Preference Optimization}
\date{} 
\author{
    Nazanin Mohammadi Sepahvand$^{1,2}$\thanks{Correspondence to: sepahvan@mila.quebec,  gkdz@google.com}
    \quad Eleni Triantafillou$^{3}$ 
    \quad Hugo Larochelle$^{2}$ 
    \quad Doina Precup$^{1,2, 3}$ \\
    \quad \quad \quad \quad \textbf{Daniel M. Roy} $^{3, 4, 5}$ 
    \quad \textbf{Gintare Karolina Dziugaite}$^{1,2, 3}$ 
    \vspace{+1.em}
    \\
    $^{1}$McGill University, Canada 
    \quad $^{2}$Mila, Canada
    \quad $^{3}$Google DeepMind 
    \quad $^{4}$University of Toronto, Canada \\
    \quad $^{5}$ Vector Institute, Canada \\
}

\begin{document}
\maketitle
\pagestyle{plain} 
\begin{abstract}
Large language models (LLMs) trained on webscale data can produce toxic outputs, raising concerns for safe deployment. Prior defenses, based on applications of DPO, NPO, and similar algorithms, reduce the likelihood of harmful continuations, but not robustly so: they are vulnerable to adversarial prompting and easily undone by fine-tuning--based relearning attacks. Indeed, research has shown that these edits to the model are superficial: linear probing reveals that harmful ``directions'' remain present in representations. To address this, we propose Representation Erasure-based Preference Optimization (\REPO), reformulating detoxification as a token-level preference problem. Using a novel objective with preference data, we force the representations of toxic continuations to converge toward their benign counterparts. Our mechanistic analysis reveals that this granular approach is critical: unlike baselines, \REPO induces deep, localized edits to toxicity-encoding neurons while preserving general model utility. Exhaustive evaluations show that \REPO achieves state-of-the-art robustness, stopping sophisticated threats---including relearning attacks and enhanced GCG jailbreaks---where existing representation- and output-based methods fail.
\end{abstract}

\section{Introduction}

LLMs trained on massive, uncurated corpora can exhibit undesirable behaviors, including memorization and regurgitation of hazardous knowledge~\citep{li2024}, the generation of toxic language~\citep{Wen23EMNLP}, and the amplification of social biases ingrained in large-scale web data~\citep{sheng2019,gehman2020}. These risks have motivated a growing set of alignment and detoxification algorithms. However, many such interventions are \emph{fragile}: while they can mitigate harmful behaviors under standard evaluations, models often remain vulnerable to jailbreak attacks that bypass safeguards and elicit harmful generations~\citep{singh2025,schwinn2024}. A prominent example is Greedy Coordinate Gradient (GCG), which appends optimized suffixes to harmful queries and achieves high jailbreak success rates across a variety of aligned models~\citep{zou2023,jia2024improved}.

Unlearning has emerged as a complementary strategy for mitigating problematic content in pretraining data, including private information and toxic language~\citep{Liu2025,Xu23unlearningsurvey}. Unlike safety training approaches that primarily suppress harmful outputs, unlearning aims to \emph{remove hazardous capabilities} from models, making them inaccessible even to adversaries with white- or black-box access~\citep{singh2025,Liu2025}. Early results suggest that unlearning can be effective against certain attacks; for example, RMU~\citep{li2024} was observed to be more resistant to linear probing of internal activations~\citep{burns2023discovering} and to classic adversarial prompting such as GCG~\citep{zou2023,ht2025rmu}. Relatedly, latent adversarial training can strengthen robustness by perturbing intermediate activations during training to suppress undesirable behaviors~\citep{sheshadri2024latent}.

\begin{figure*}[t]
    \centering
    \includegraphics[width=.9\linewidth]{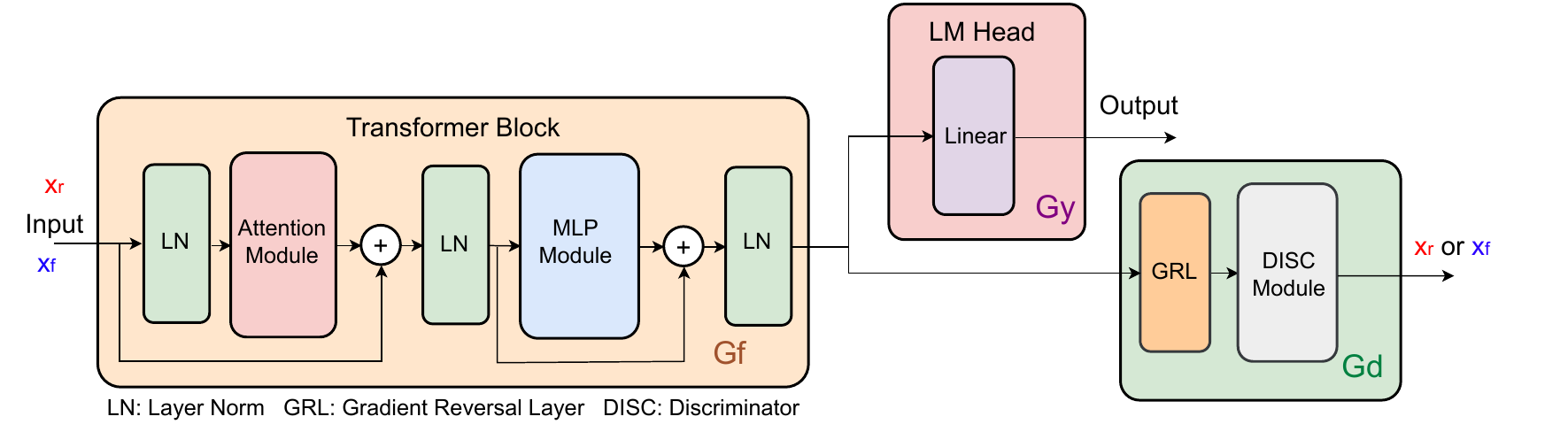}
    \caption{A schematic representation of \REPO. Its regressor can be attached to any transformer block $M$ targeted for unlearning; here, $M$ is taken as the final transformer block before the linear unembedding layer. For each prompt, the \textcolor{red}{retain (nontoxic) continuation $x_r$} and the \textcolor{Blue}{forget (toxic) continuation $x_f$} are fed into the network, and the discriminator is trained to distinguish between toxic and nontoxic inputs.}
    \label{fig:schematic}
\end{figure*}

However, these unlearning methods have been shown to fail against more adaptive jailbreaks~\citep{lucki2409adversarial,singh2025,hu2024jogging}. Among the most effective recovery strategies are \emph{relearning} attacks, which recover supposedly removed capabilities through lightweight fine-tuning on as few as ten unrelated examples~\citep{hu2024jogging}, and \emph{enhanced} variants of GCG that substantially improve attack success against methods such as RMU and NPO with only small modifications to the original loss~\citep{lucki2409adversarial}. These results suggest that reducing the likelihood of harmful outputs is often easier than removing the \emph{internal representational affordances} that enable harmful generation and can be reactivated under distribution shift or retraining.

Motivated by these vulnerabilities, recent work has increasingly explored representation-based approaches that intervene directly on hidden representations rather than only on model outputs. Embedding-based unlearning has been shown to be more resilient to paraphrasing attacks, preventing forgotten knowledge from resurfacing under semantic variations of prompts~\citep{spohn2025}. Mechanistic localization to factual recall pathways can likewise improve robustness against relearning by preventing capabilities from being restored through lightweight fine-tuning~\citep{guo2024}. Representation-level interventions also resist membership inference and inversion attacks, offering stronger privacy guarantees for forgotten data~\citep{hu2025}. Overall, these findings suggest that targeting hidden features can enable more durable forgetting and improved resistance to knowledge recovery compared to output-level or gradient-based approaches~\citep{muhamed2025,jung2025,wang2025}.

A natural formalization of this idea is \emph{representation erasure}: remove decodable information about an undesirable attribute from internal states, so that downstream computation cannot reliably act on it. In classification, representation-erasure methods such as SURE~\citep{sepahvand2025sure} provide evidence that adversarial invariance objectives can yield robust forgetting. Translating representation erasure to generative LLMs, however, requires a fundamental shift in approach. Unlike classification---where one can ``scrub'' a single representation vector---detoxifying an autoregressive model requires controlling toxic features \emph{at the token level} within a continuous generation stream.

This paper proposes \emph{Representation Erasure-based Preference Optimization} (\REPO), which adapts representation erasure to generative detoxification using \emph{pairwise} supervision. For each prompt $x_p$, we assume a preferred \emph{retain} continuation $x_r$ (nontoxic) and a dispreferred \emph{forget} continuation $x_f$ (toxic). \REPO combines two objectives: (i) a token-level anchoring loss that matches the edited model to a frozen reference model on retain continuations, preserving benign behavior, and (ii) a token-granular adversarial objective that makes retain and forget token representations indistinguishable, aligning toxic continuations toward their benign counterparts and removing the features that distinguish harmful tokens. This design differs from classification-oriented erasure in two key ways: it is prompt-conditioned via paired retain/forget continuations, and it operates at token granularity to match the autoregressive decoding process~\citep{sepahvand2025sure}.

Like DPO \citep{lee2024mechanistic}, \REPO leverages pairwise supervision (preferred vs.\ dispreferred continuations). However, while DPO enforces preferences in output space (likelihoods), \REPO enforces them in representation space, removing the internal features that distinguish toxic sequences. This renders \REPO robust against adaptive prompting and lightweight fine-tuning attacks.


Our contributions are as follows:
\begin{itemize}[noitemsep, topsep=0pt, leftmargin=*]
\item We introduce \REPO, a pairwise, token-level representation-erasure objective for detoxifying LLMs that couples reference anchoring on benign text with adversarial invariance between retain and forget representations.
\item We evaluate \REPO under adaptive recovery settings, including relearning and enhanced jailbreak attacks~\citep{hu2024jogging,lucki2409adversarial}, 
demonstrating superior detoxification and robustness compared to state-of-the-art, while preserving utility.
\item We provide mechanistic analyses and ablations showing that  the representation-level objective is responsible for the edits deep into the network, 
while the token-level granularity is critical for the localized precision.


\end{itemize}

\section{\REPO for Detoxifying LLMs} 
\label{sec:ourmethod}

We assume a paired dataset of triples
$$\textstyle
\smash{\mathcal{D}=\{(x_p^{(i)}, x_r^{(i)}, x_f^{(i)})\}_{i=1}^N,}
$$
where each prompt $x_p$ is paired with a \emph{retain} continuation $x_r$ (non-toxic) and a \emph{forget} continuation $x_f$ (toxic). We write the corresponding full sequences as 
$
s_r = [x_p; x_r]$ and $ s_f = [x_p; x_f]$,
then assign a domain label $d(s)\in\{0,1\}$ with $d(s_r)=0$ and $d(s_f)=1$.

Our goal is to edit the model to (i) preserve the original model's behavior on retain data, and (ii) remove representational features that enable toxic generation on forget data.

\subsection{Model components}
Let the LLM be decomposed into (i) a transformer feature extractor $G_f(\cdot;\theta_f)$ that maps an input sequence $s$ to token representations $\{h_t(s)\}_{t=1}^{|s|}$, and (ii) an LM head $G_y(\cdot;\theta_y)$ that maps $h_t(s)$ to logits $z_t(s)$ and next-token distributions
\[\textstyle
\smash{\pi_\theta(\cdot \mid s_{\le t})=
\pi_\theta^t(s),
\qquad \text{where } \theta=(\theta_f,\theta_y).}
\]
We also define a frozen reference model $\theta^{\mathrm{ref}}$ (the original pretrained parameters), used only for anchoring.

To implement representation erasure, we attach a small discriminator (e.g., a two-layer MLP) $G_d(\cdot;\theta_d)$ to token representations at a chosen transformer layer $\ell$ (in our experiments, the final transformer block before unembedding). The discriminator is connected through a \emph{gradient reversal layer} $R(\cdot)$ \citep{ganin2016dann}, which is the identity on the forward pass and multiplies gradients by $-1$ on the backward pass. The discriminator outputs a domain probability
$$
q_t(s) = \smash{G_d\!\big(R(h_t^{(\ell)}(s));\theta_d\big)\in(0,1),}
$$
interpreted as $q_t(s)\approx \Pr(d(s)=1 \mid h_t^{(\ell)}(s))$.

\paragraph{Why representation erasure affects generation (informal rationale).}
In an autoregressive LM, the next-token distribution is a function of the hidden representation used by the LM head. In standard transformer LMs, logits are linear in the final representation, $z_t = W h_t + b$. If, for a fixed prompt, representations along toxic continuations are driven to match those along benign continuations at the layer feeding the head, then the logits—and therefore the next-token distributions—match as well, preventing the model from reliably continuing along a toxic trajectory beyond what the benign trajectory would produce. REPO approximates this matching by making retain and forget representations \emph{indistinguishable} to a discriminator, while explicitly anchoring retain behavior to prevent trivial collapse.

\subsection{REPO objective}

\paragraph{Retain anchoring loss (token-level KL).}
To preserve benign behavior, we minimize a token-wise KL divergence between the edited model and the frozen reference model on retain sequences:
\begin{equation*}
\begin{aligned}
\mathcal{L}_{\text{retain}}(\theta)
&:= \mathbb{E}_{s_r \sim \mathcal{D}_r}\!\bigg[
\frac{1}{|s_r|} \smash{\sum_{t=1}^{|s_r|}
\KL \Big(\pi_{\theta^{\mathrm{ref}}}^t(s_r)\,\big\|\,\pi_{\theta}^t(s_r)\Big)}
\bigg],
\end{aligned}
\end{equation*}

where $\mathcal{D}_r$ denotes the retain distribution induced by $\mathcal{D}$. (The key property is token-level anchoring to the reference distribution across the entire prompt+retain sequence; either consistent KL direction may be used.)

\paragraph{Representation erasure loss (token-level domain adversarial).}
We train the discriminator to predict whether a token representation came from a retain or forget sequence, and simultaneously train the LLM to \emph{fool} the discriminator. Using binary cross-entropy (BCE),
\begin{equation*}
\mathcal{L}_{\text{dom}}(\theta_f,\theta_d)
=
\mathbb{E}_{s\sim \mathcal{D}_r\cup \mathcal{D}_f}
\bigg[
\frac{1}{|s|}
\smash{\sum_{t=1}^{|s|}
\mathrm{BCE}\big(q_t(s),\, d(s)\big)}
\bigg],
\end{equation*}
where $\mathcal{D}_f$ denotes the forget distribution induced by $\mathcal{D}$.

\paragraph{Minimax form and implemented training loss.}
The intended optimization is
\begin{align*}
\smash{\min_{\mathclap{\theta_f,\theta_y}} \quad}
& \alpha\,\mathcal{L}_{\text{retain}}(\theta)
\;-\;
(1-\alpha)\,\mathcal{L}_{\text{dom}}(\theta_f,\theta_d),
\\
\smash{\min_{\theta_d}\quad}
& \mathcal{L}_{\text{dom}}(\theta_f,\theta_d),
\end{align*}
where $\alpha\in[0,1]$ trades off utility preservation and erasure pressure. In practice, we compute $\mathcal{L}_{\text{dom}}$ normally but connect $G_d$ to $G_f$ through $R(\cdot)$, which flips the gradient sign flowing into $\theta_f$ and implements the minimax without explicit alternating maximization steps (See Algorithm~\ref{alg:repo} for a formal summary of the method described above.)

\subsection{Modeling choices}
\paragraph{Token-level granularity.}
A sequence-level discriminator (e.g., pooling representations before classification) can encourage coarse alignment while allowing toxic information to remain localized in specific positions. REPO instead applies the discriminator \emph{per token} and averages over tokens. This targets the parts of the computation graph that encode toxic tokens and their immediate causal footprint, while the retain KL discourages broad degradation of language modeling behavior.

\paragraph{Why ``preference optimization'', and how REPO differs from DPO/NPO.}
We call REPO a preference optimization method because the supervision is pairwise: for each prompt, we are given a preferred continuation $x_r$ and a dispreferred continuation $x_f$. DPO/NPO use this pairing to shift \emph{output-space} likelihoods, typically by increasing the relative log-probability of $x_r$ over $x_f$ (often with an implicit or explicit regularization toward a reference model). REPO uses the same pairing differently: it (i) anchors the model to the reference on preferred text via token-wise KL, and (ii) uses the rejected text to drive \emph{representation-level} erasure through a domain-adversarial objective. This is designed to address recoverability: rather than only making toxic outputs less likely under current decoding, REPO removes decodable internal features that distinguish toxic continuations from benign ones under the same prompt. From that perspective, REPO can be viewed as an unlearning algorithm, unlike prior preference optimization methods that do not erase the knowledge of the dispreferred continuations from internal features.

\section{Evaluation Metrics}

\begin{figure*}[t]
  \centering
  \includegraphics[width=.9\linewidth]{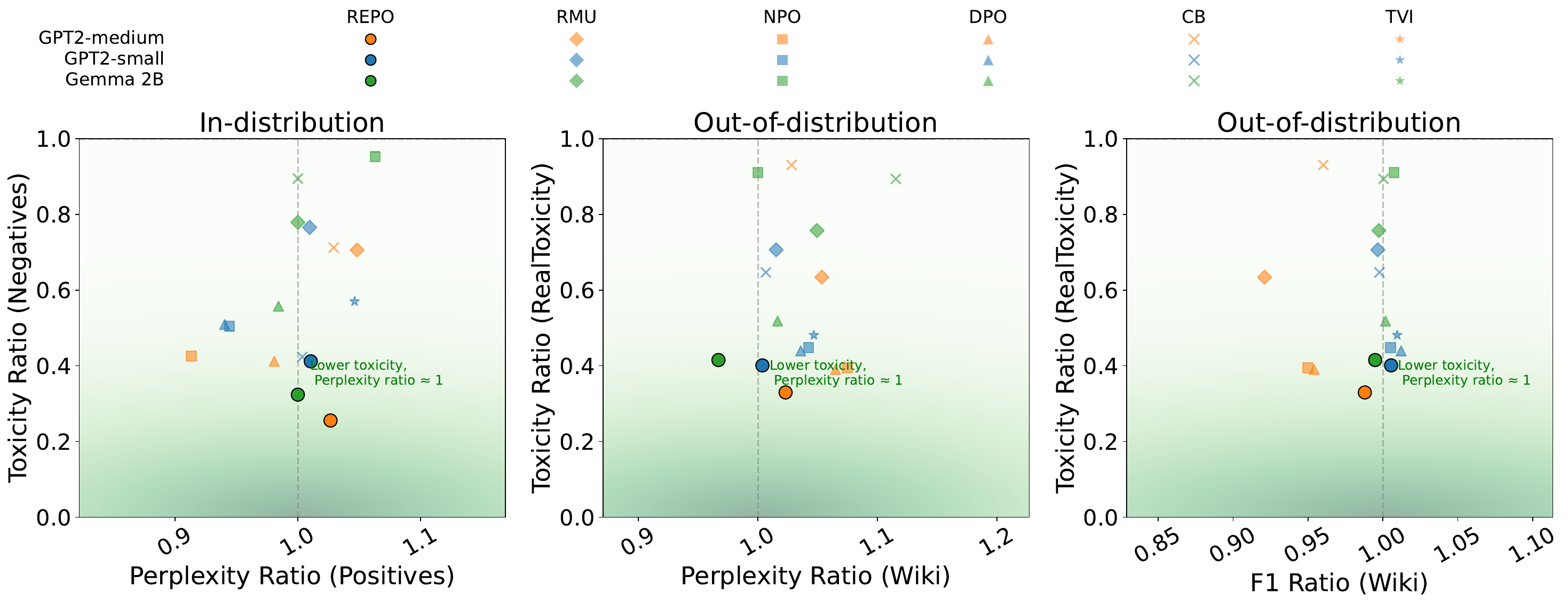}
  \caption{Detoxified models vs reference. 
\textbf{(Left)} Perplexity vs.\ toxicity ratios on PairToxicity (in-distribution); 
\textbf{(Middle)} Perplexity vs.\ toxicity ratios on WikiText/RealToxicity (OOD); 
\textbf{(Right)} F\textsubscript{1} ratio on WikiText vs.\ OOD toxicity. 
Each point is a model–method pair. 
The green gradient highlights lower toxicity and ratios near 1, darkest at the ideal point \((x\!=\!1,\;y\!=\!0)\). 
Dashed gray lines mark ratio = 1 for easy comparison to the reference.}
  \label{fig:metrics}
\end{figure*}

We evaluate our approach along two complementary dimensions: (i) its effectiveness in removing toxic behaviors while preserving general capabilities; this is often referred to as unlearning-utility trade-off, and (ii) its robustness against adaptive attacks aimed at reactivating toxic behaviors. Below we describe the metrics used in each case.

\subsection{Effectiveness} 


\par{\textbf{Toxicity Score.}} Following prior work~\citep{geva2022, lee2024mechanistic}, we evaluate toxicity using the Perspective API, an automated tool for toxicity detection that estimates the probability a continuation would be perceived as toxic.

\par{\textbf{Utility.}}
We evaluate utility using Perplexity and $F_1$ score on WikiText-2 \citep{merity2016}. Perplexity proxies divergence from the pretrained distribution, while $F_1$ measures overlap with ground-truth continuations. (See \cref{app:utility}.)

\subsection{Robustness} 
A key challenge in unlearning is robustness: toxic behavior may disappear, only for an adversary to recover it. We consider three attack strategies studied in the unlearning literature~\citep{wang2025invariance, lucki2409adversarial, hu2024jogging}: relearning, orthogonalization, and enhanced GCG. For the latter two, model weights remain frozen and only inference-time manipulations are applied, whereas relearning modifies the model via fine-tuning. Attack effectiveness is quantified by comparing the toxicity of generations from the unlearned model before and after the attack.

\textbf{Relearning Attack.} Prior studies have shown that fine-tuning can easily reverse alignment or unlearning, even when the fine-tuning data is small or consists of datasets with low mutual information with the forget set~\citep{wang2025invariance, lucki2409adversarial, hu2024jogging, siddiqui2025dormant}. As in prior work~\citep{lucki2409adversarial}, we fine-tune unlearned models under two configurations: (i) on 10 examples from the forget set, and (ii) on 1000 examples from the retain set. The former evaluates recovery under minimal direct exposure, while the latter tests recovery using data with low mutual information with the forgotten knowledge.

\begin{figure}[t]
    \centering
    \vspace{-2em}
    \includegraphics[width=.7\linewidth]{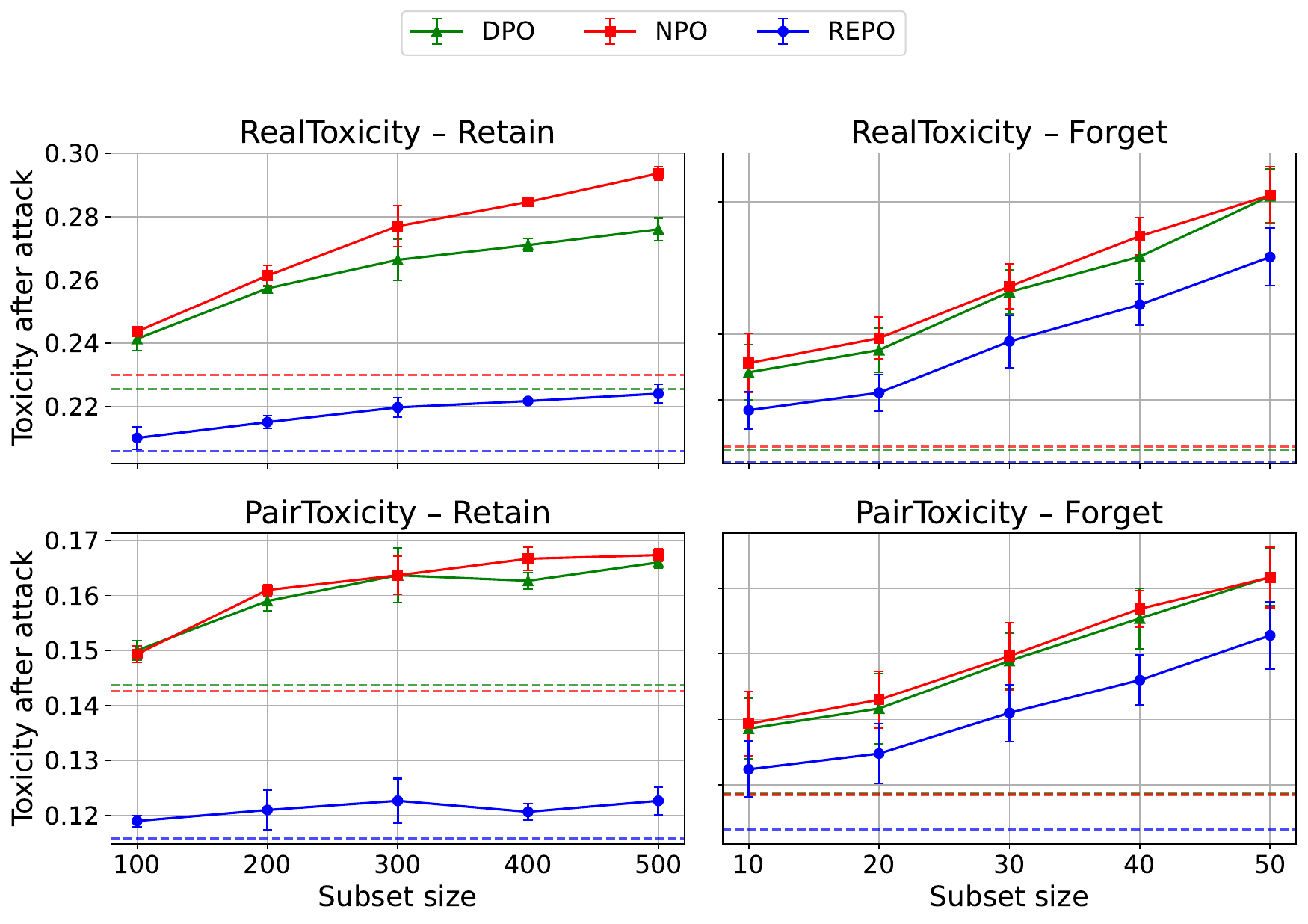}
    \caption{Average toxicity after the \textit{Relearning Attack} for different subset sizes across methods on GPT2-small. 
    \textbf{(Top)} OOD toxicity (RealToxicity); \textbf{(Bottom)} In-distribution toxicity (pairwise set). 
    Dashed horizontal lines indicate each method's baseline toxicity before the attack.}
    \label{fig:relearn_attack}
    \vspace{-1em}
\end{figure}

\textbf{Orthogonalization Attack.} Previous work demonstrated that safety refusals can often be attributed to a direction in activation space~\citep{arditi2024refusal}. \citet{lucki2409adversarial} extended this idea to the unlearning setting. Following their approach, we compute an \textit{unlearned direction} for each transformer block as the difference in mean activations between the reference and unlearned models on the forget set~\citep{lucki2409adversarial, belrose2023diffinmeans}. At inference time, this direction is projected out of the hidden representations, thereby removing the offset introduced by unlearning and restoring toxic capabilities.

\textbf{Enhanced GCG Attack.} GCG attacks have been reported ineffective against representation-based unlearning methods such as RMU~\citep{li2024, lucki2409adversarial}. We thus adopt an enhanced variant that specifically targets unlearning defenses~\citep{lucki2409adversarial}. Rather than minimizing the standard attacker loss toward generating a fixed affirmative target string~\citep{zou2023}, the attack leverages the reference model as a malicious teacher. Concretely, adversarial prefixes are optimized with a distillation loss that aligns the unlearned model's hidden representations at selected layers with those of the reference model~\citep{thompson2024flrt}. This adaptation enables recovery of harmful behaviors that classic GCG cannot elicit.

\section{Experimental Details}
\label{sec:expdetails}

\paragraph{Data and Models.} Our evaluation relies on three datasets serving complementary purposes: a pairwise toxicity dataset for unlearning, PairToxicity, \citep{lee2024mechanistic}, WikiText-2 \citep{merity2016} for measuring generation quality, and RealToxicityPrompts \citep{gehman2020} for assessing OOD toxicity. 
We evaluate our approach on GPT-2 Small, GPT-2 Medium \citep{radford2019language} and Gemma 2B (base) \citep{team2024gemma}. See \cref{app:details} for further details.

\begin{figure}[t]
  \centering
  \includegraphics[width=0.95\linewidth]{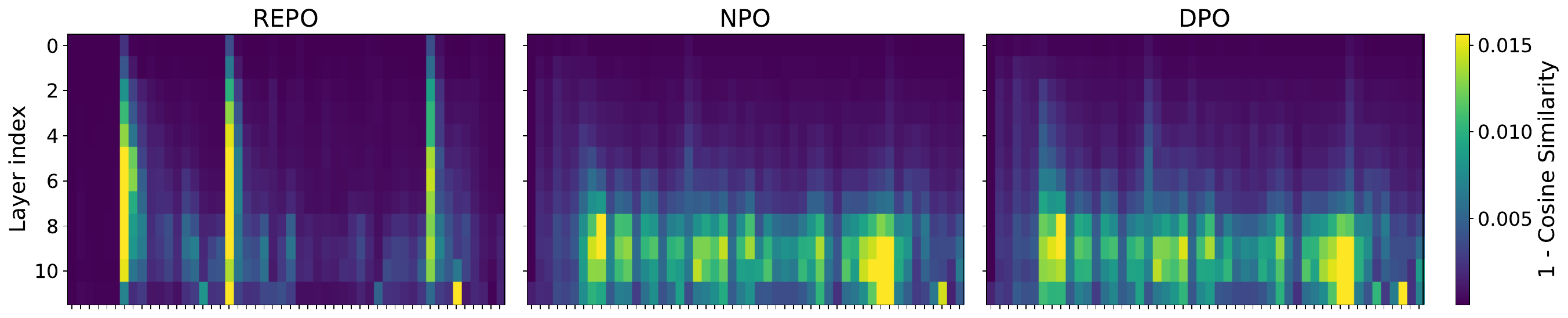}\\[-.5em] 
  \includegraphics[width=.95\linewidth]{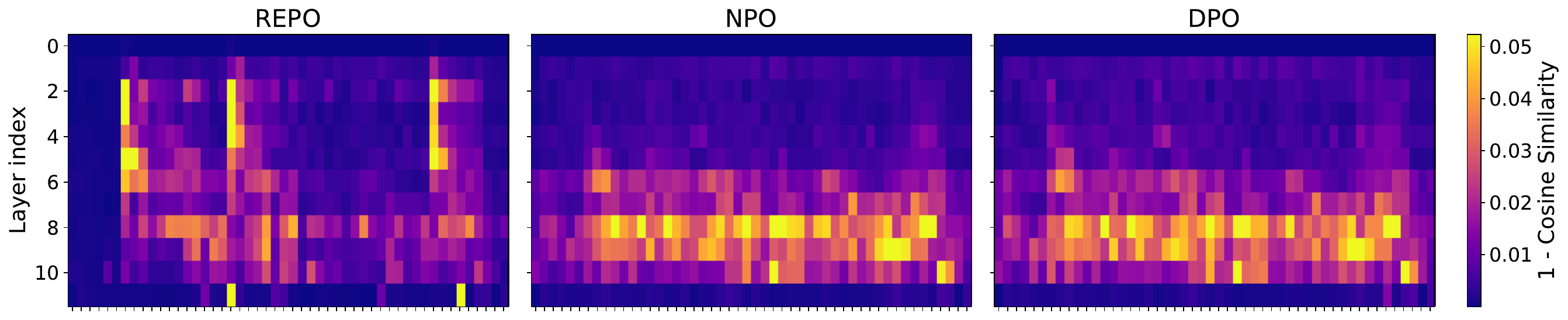}\\[-.5em] 
\includegraphics[width=.95\linewidth]{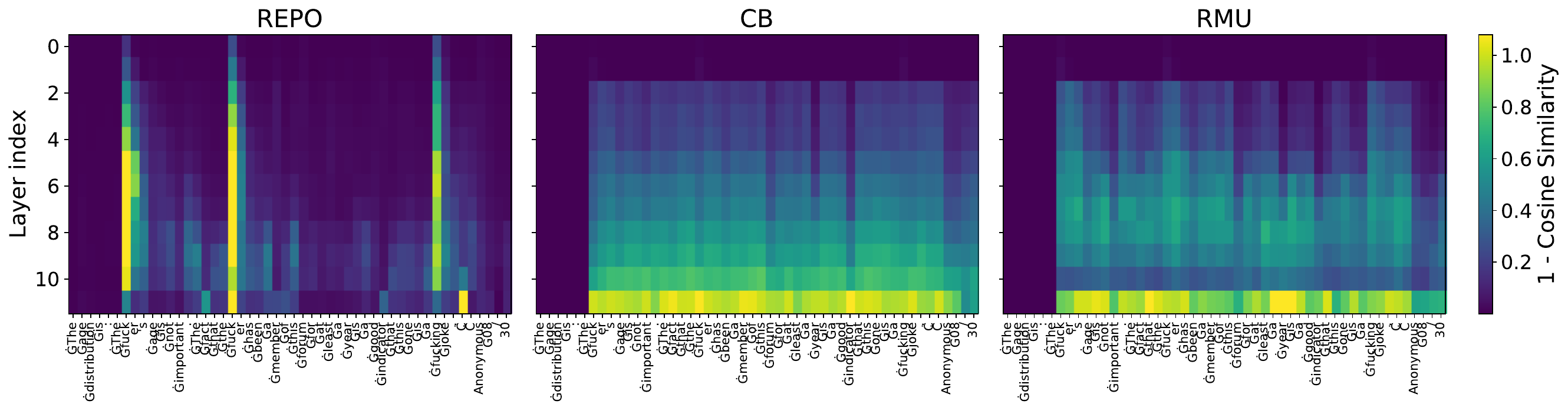}
  \caption{Layer–token distance heatmaps for different methods on a sample prompt. Columns show (left to right) \REPO, NPO, and DPO (top two rows), and \REPO, CB, and RMU (bottom row).
\textbf{Top:} $1-\cos$ similarity between unlearned and reference hidden states across GPT-2 small layers (y-axis) and tokens (x-axis); darker indicates higher similarity.
\textbf{Middle:} $1-\cos$ similarity between attention submodule outputs (before residual addition) of the unlearned and reference models.
\textbf{Bottom:} Same as the top row, but for representation-based methods.
  }
  \label{fig:layer_token_distance}
  \vspace{-1em}
\end{figure}


\paragraph{Baselines.} We compare \REPO against two main families of alignment methods: \textit{steering-based} and \textit{fine-tuning–based}, the latter being further subdivided into \textit{representation-} and \textit{output-based}. 

\textit{Steering-based methods} act directly on hidden representations at inference time, modifying activations to suppress toxic behaviors without retraining, or even finetuning, the model. As a representative baseline, we adopt Toxic Vector Intervention \citep[TVI;][]{lee2024mechanistic}, which operates by subtracting identified toxic vectors from the model's activations during generation---a lightweight steering-style approach.

\textit{Fine-tuning–based methods} explicitly retrain the model to remove undesired behaviors. We further divide them into two categories. \textit{Output-space methods} operate directly on the model's output probabilities. Among them, we include preference-based objectives such as Direct Preference Optimization (DPO) and Negative Preference Optimization \citep[NPO;][]{wang2025invariance, lucki2409adversarial}, which fine-tune the model to increase the likelihood of preferred continuations and decrease the relative likelihood of undesired ones. \textit{Representation-space methods} operate on hidden activations. Examples include Representation Misdirection for Unlearning \citep[RMU;][]{li2024, ht2025rmu, kadhe2024rmu} and Circuit Breakers \citep[CB;][]{zou2024circuitbreakers}, which were originally proposed to erase hazardous knowledge but here are adapted to the detoxification setting. While RMU maps toxic directions in representation space to random directions, CB severs identified causal pathways associated with harmful behavior.

\begin{figure}[t]
    \centering
    \includegraphics[width=1\linewidth]{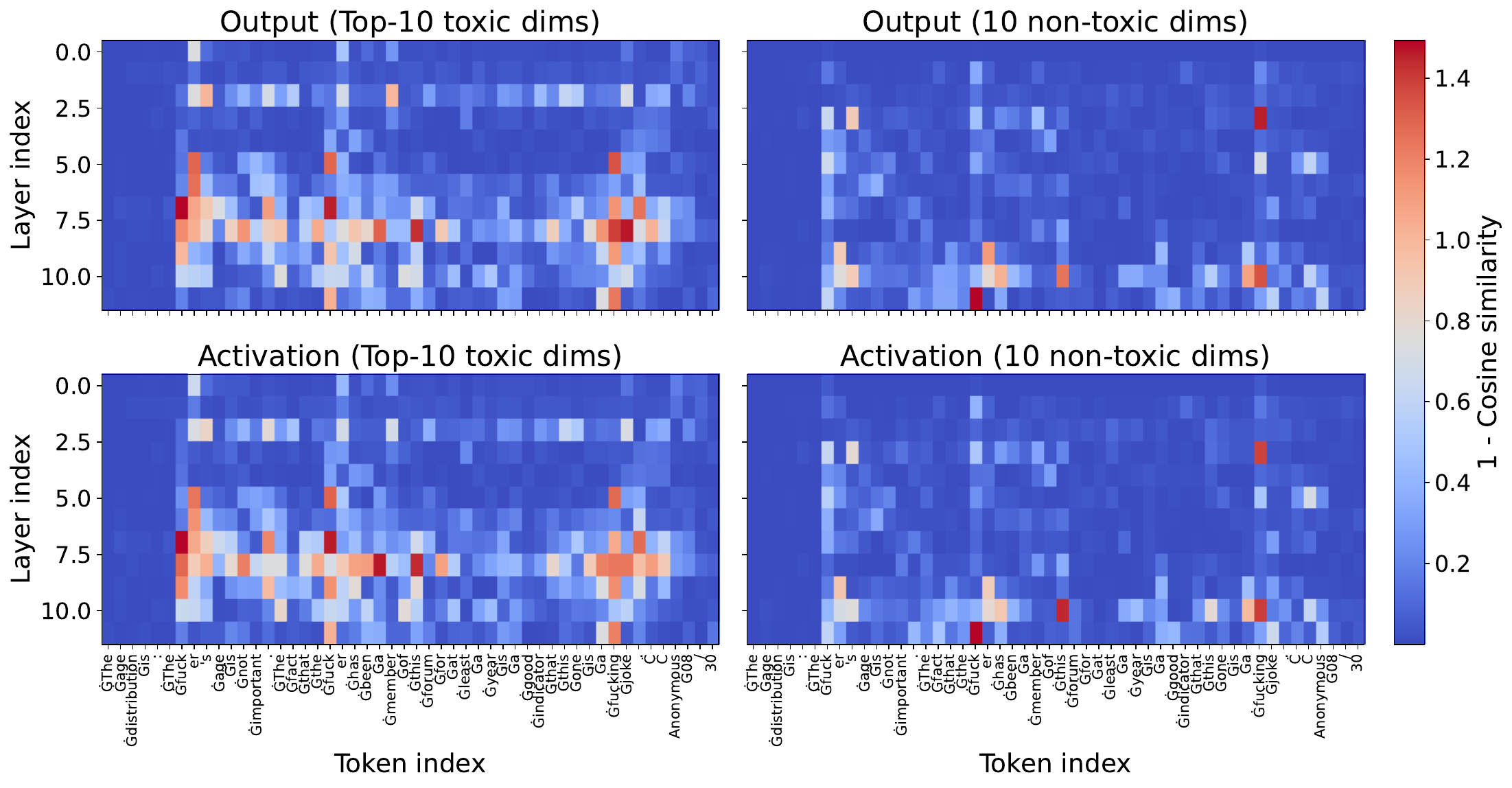}
    \caption{Layer--token residual-stream drift ($1-$cosine similarity) between the reference and \REPO models for the same negative prompt. \textbf{Top:} Differences in residual contributions (post-activation keys multiplied by value vectors). \textbf{Bottom:} Differences in key activations. Within each row, \textbf{Left} shows the top-10 toxic dimensions (most aligned with $W_{\text{toxic}}$) and \textbf{Right} shows 10 non-toxic dimensions. Rows correspond to GPT-2 Small layers and columns to prompt tokens; darker colors indicate greater similarity and yellow larger drift.}
    \label{fig:heatmap_toxic}
\end{figure}


\begin{figure}[t]
\centering

\begin{minipage}{0.49\linewidth}
  \centering
  \includegraphics[width=\linewidth]{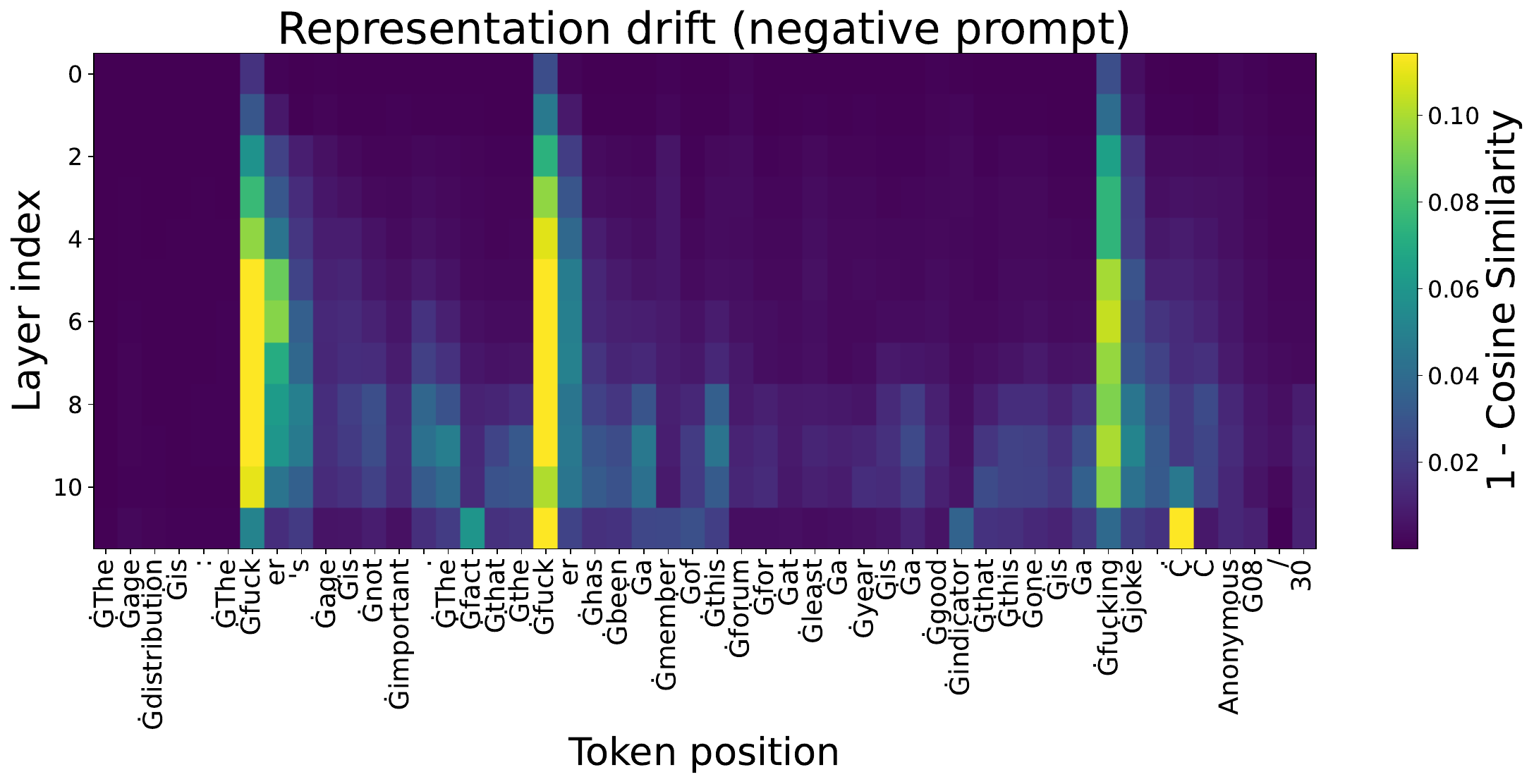}
\end{minipage}
\hfill
\begin{minipage}{0.49\linewidth}
  \centering
  \includegraphics[width=\linewidth]{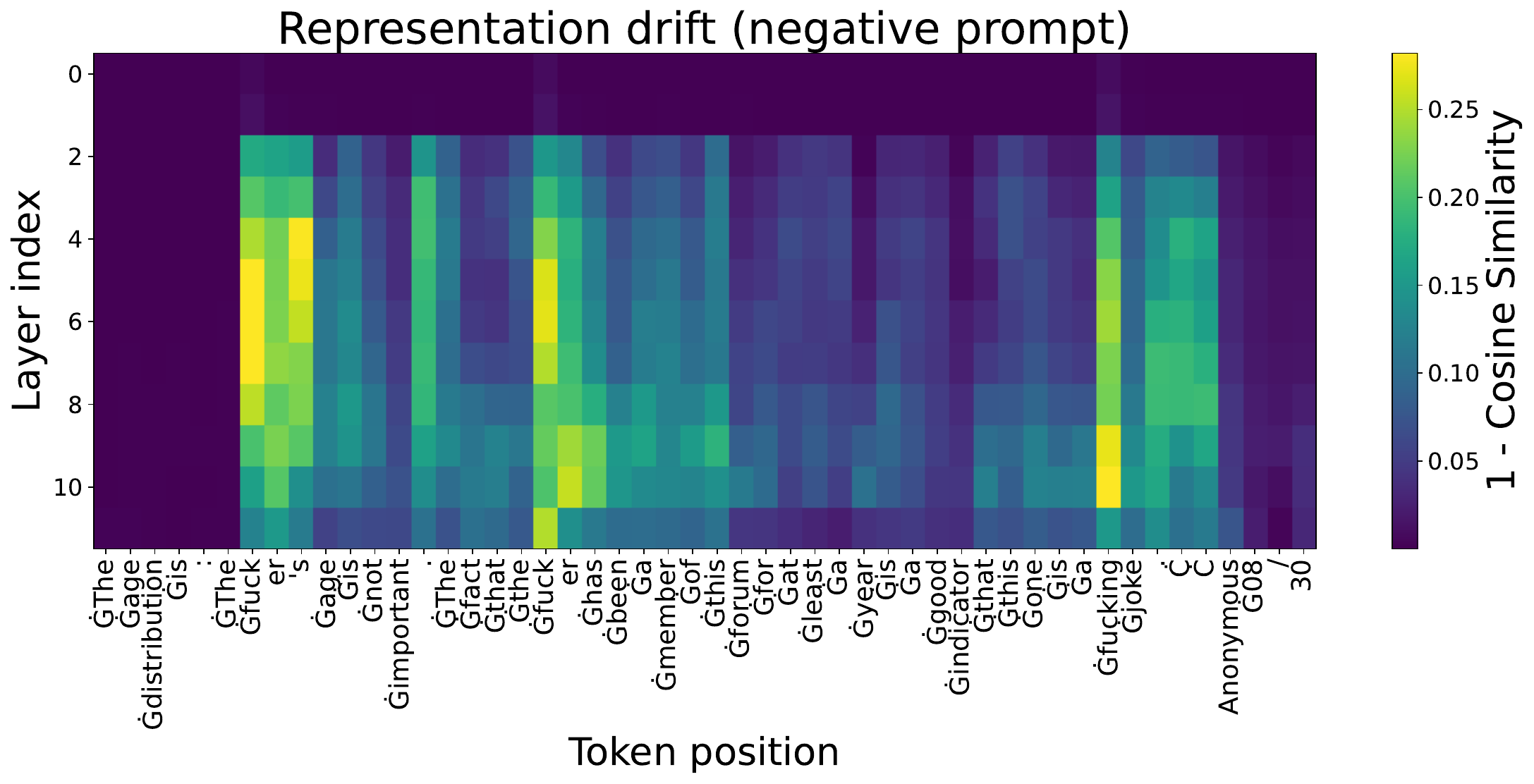}
\end{minipage}

\caption{Layer–token representation drift (\(1-\)cosine similarity) for the same negative prompt under two discriminator input strategies in \REPO: \textbf{Left} — individual tokens; \textbf{Right} — non-overlapping averaged segments. Darker colours indicate greater similarity, yellow larger drift.}
\label{fig:heatmap_SURE_token_vs_avg}
\end{figure}


\begin{table}[t]
\caption{%
Robustness of unlearning methods on GPT-2 (Medium) and Gemma-2B across PairToxicity and RealToxicity. Rows denote evaluation setups and columns denote methods. Each cell reports post-attack toxicity, with parentheses indicating toxicity immediately after unlearning (pre-attack). Baseline toxicity before unlearning: GPT-2—0.281 (PairToxicity), 0.513 (RealToxicity); Gemma-2B—0.208 (PairToxicity), 0.486 (RealToxicity).%
}
\label{tab:eval}
\centering
\begin{tabular}{llcccccc}
\hline
 & & \textbf{\REPO} & \textbf{NPO} & \textbf{DPO} & \textbf{RMU} & \textbf{CB} \\
\hline
\multirow{6}{*}{\textbf{GPT-2}}
  & Relearning Forget (PairToxicity) & \bf .169(.116) & .202(.143) & .200(.144) & .253(.215) & .438(.120) \\
  & Relearning Retain (PairToxicity) & \bf .119(.116) & .148(.143) & .148(.144) & .204(.215) & .124(.120) \\
  & Relearning Forget (RealToxicity) & \bf .294(.206) & .377(.230) & .357(.224) & .463(.363) & .678(.332) \\
  & Relearning Retain (RealToxicity) & \bf .207(.206) & .245(.230) & .237(.224) & .362(.363) & .314(.332) \\
  & Enhanced-GCG (RealToxicity) & \bf .208(.206) & .347(.230) & .660(.224) & .389(.363) & .393(.332) \\
  & Orthogonalization (RealToxicity) & \bf .308(.206) & .335(.230) & .315(.224) & .525(.363) & .335(.332) \\
\hline
\multirow{6}{*}{\textbf{Gemma-2B}}
  & Relearning Forget (PairToxicity)  & \bf .108(.083) & .255(.247) & .169(.146)  & .329(.206) & .161(.160) \\
  & Relearning Retain (PairToxicity)  & \bf .089(.083) & .249(.247) & .169(.146) & .212(.206) & .162(.160) \\
  & Relearning Forget (RealToxicity)  & \bf .257(.215) & .461(.439) & .304(.244) & .579(.356) & .402(.412) \\
  & Relearning Retain (RealToxicity)  & \bf .216(.215) & .453(.439) & .304(.244) & .344(.356) & .421(.412) \\
  & Enhanced-GCG (RealToxicity) & \bf .217(.215) & .472(.439) & .269(.244) & .358(.356) & .428(.412) \\
  & Orthogonalization (RealToxicity) & \bf .217(.215) & .442(.439) & .248(.244) & .357(.356) & .415(.412) \\
\hline
\end{tabular}
\end{table}

\section{Performance Evaluation}

Our initial analysis focuses on several key performance aspects. As standard, we first examine the trade-offs between unlearning quality and model utility. This evaluation covers both in-distribution performance on the PairToxicity dataset used for unlearning, and out-of-distribution (OOD) generalization on RealToxicityPrompts and WikiText-2. Here we see that \REPO has superior performance across the board. In addition, more qualitative results are presented in \ref{app:qualresults}. We then evaluate robustness to various attacks, observing \REPO's competitive performance. 

\paragraph{Mitigating Toxicity vs Preserving Utility}
\cref{fig:metrics} reports results on the in-distribution dataset (PairToxicity). For GPT2-Small, \REPO achieves the lowest toxicity on negative (forget) samples (0.0961), substantially outperforming NPO (0.1392), DPO (0.1506), and RMU (0.1527). Importantly, toxicity on positive (retain) samples remains comparable, showing that nontoxic generations are preserved. Perplexity results indicate that \REPO increases uncertainty on toxic continuations (70.8 vs.\ 18.1 for the reference model) while leaving perplexity on retain samples largely unchanged, consistent with its goal of erasing toxic information. These results demonstrate that \REPO effectively targets toxic continuations without impairing general language modeling, with the same trend observed for GPT2-Medium and Gemma-2B.

As shown in \cref{fig:metrics}, \REPO achieves state-of-the-art OOD performance. On GPT2-Small, it yields the lowest RealToxicityPrompts score ($0.21$), outperforming the next-best NPO ($0.24$), while virtually matching the reference model's utility (WikiText PPL $23.6$). This superior unlearning-utility trade-off generalizes across GPT2-Medium and Gemma-2B.



While perplexity and $F_1$ score assess fluency and generation quality, they do not capture performance on downstream tasks. To evaluate this, we benchmarked all unlearned model variants alongside the reference model (REF) on MMLU~\citep{hendrycks2020measuring} using Gemma-2B. MMLU comprises a diverse set of language understanding tasks across multiple domains, each posed as a multiple-choice problem. We focus on Gemma-2B because smaller models exhibit near-random baseline accuracy, making downstream evaluation unreliable. As shown in Table~\ref{tab:mmlu}, all detoxification methods achieve accuracy comparable to REF, indicating that toxicity mitigation does not degrade downstream task performance.

\begin{table}[h]
\centering
\caption{MMLU accuracy on Gemma-2B for the reference model (REF) and all detoxification methods. Values report mean accuracy averaged across all MMLU tasks.}
\label{tab:mmlu}
\begin{tabular}{l c c c c c c}
\toprule
Method & REF & REPO & DPO & NPO & CB & RMU \\
\midrule
Accuracy & 0.418 & 0.422 & 0.422 & 0.423 & 0.418 & 0.411 \\
\bottomrule
\end{tabular}
\end{table}

\noindent\textbf{Robustness to Attacks.} 
Table~\ref{tab:eval} evaluates robustness under adversarial attacks for GPT2-Medium and Gemma-2B. We consider three attack types: relearning, where the forget set is reintroduced via lightweight fine-tuning; enhanced GCG, which increases the success of adversarial prompting; and orthogonalization. \REPO consistently outperforms all baselines across attacks. For instance, for GPT2-Medium on the PairToxicity dataset under relearning, \REPO achieves lower toxicity on retain samples (0.207 vs.\ 0.245 for NPO and 0.237 for DPO) and on forget samples (0.119 vs.\ 0.148 for both NPO and DPO). Against enhanced-GCG, \REPO again achieves the lowest toxicity (0.208 vs.\ 0.389 for RMU and 0.347 for NPO), and the same trend holds for orthogonalization. For the relearning attack on GPT2-Small, \cref{fig:relearn_attack} shows that \REPO maintains a consistent advantage over DPO and NPO across different numbers of forget and retain samples. Together these results show that \REPO resists the recovery of toxic behaviors even under stronger adversarial settings.

\section{The Effects on Representations and Weights}
The analysis in this section is focused on studying the mechanisms behind \REPO's performance. We demonstrate that \REPO has larger magnitude weight edits (\cref{app:weightspace}), but these edits result in more localized edits on conditional distributions of toxic words, and affect representations deeper in the network. Building on the analysis by \citet{lee2024mechanistic}, we then inspect the changes in value and key vectors, observing that the biggest shift happens in dimensions most \emph{and least} aligned  with toxic directions. Our ablations reveal that these differences between \REPO and other methods are due to two key algorithm design choices: (1) edits on the representations instead of output, resulting in bigger changes deeper in the network, and (2) \REPO's optimization objective being at a token-level granularity -- this ensures more localized shifts on the toxic word distribution.

\paragraph{Changes in the intermediate states.}

In \cref{app:weightspace} we show that \REPO makes larger edits in weight space. Having observed that, we now examine how these changes affect the model's intermediate representations. \cref{fig:layer_token_distance} visualizes this by plotting the representational drift (1-cosine similarity) between the unlearned and reference models' hidden states across all layers for a sample toxic continuation\fTBD{ET: one might ask: is this one cherry picked? could we show a few more examples in the appendix and/or one that averages over all toxic continuations? (would that one be too messy?)}. The heatmaps for \REPO show that modifications are highly localized. Significant drift is concentrated in the network's deeper layers, and is confined almost exclusively to the columns corresponding to the toxic tokens, while the representations for adjacent tokens show minimal change. In stark contrast, DPO and NPO induce more diffuse, lower-magnitude changes that are spread across a broader set of tokens and layers. 
This analysis provides an intuition for \REPO's good utility-unlearning trade-off: it achieves effective unlearning
by making targeted modifications to the representations of specific toxic inputs while preserving the integrity of non-toxic ones.

\section{Ablations of Algorithmic Components}

We conduct a series of ablations to dissect \REPO's design and identify the sources of its effectiveness: representation-space edits, and token-level objective.
We then provide evidence that \REPO more aggressively targets the specific neurons most aligned with toxicity compared to baselines.

\vspace*{-0.5em}

\paragraph{Changing the token-level objective.}
To isolate \REPO's components responsible for the localized edits, we conduct an ablation study on the granularity of \REPO's adversarial objective. We compare our standard approach, where the discriminator evaluates each token's representation individually, with a variant where representations are averaged over non-overlapping segments before being passed to the discriminator.
The results are visualized in \cref{fig:heatmap_SURE_token_vs_avg}. The top row, showing the standard token-level objective, exhibits the highly localized representational drift previously discussed. In contrast, the bottom row shows that using averaged segments causes this localization to vanish. The representational drift becomes diffuse, spreading across multiple tokens rather than being confined to specific ones. This diffusion in representation space correlates with a degradation in unlearning performance, yielding a worse utility-unlearning trade-off. This ablation provides strong evidence that the token-level granularity of \REPO's adversarial loss is a key mechanism responsible for the precision of its edits, which in turn contributes to its strong performance.

\begin{figure}[t]
    \centering
    \includegraphics[width=.75\linewidth]{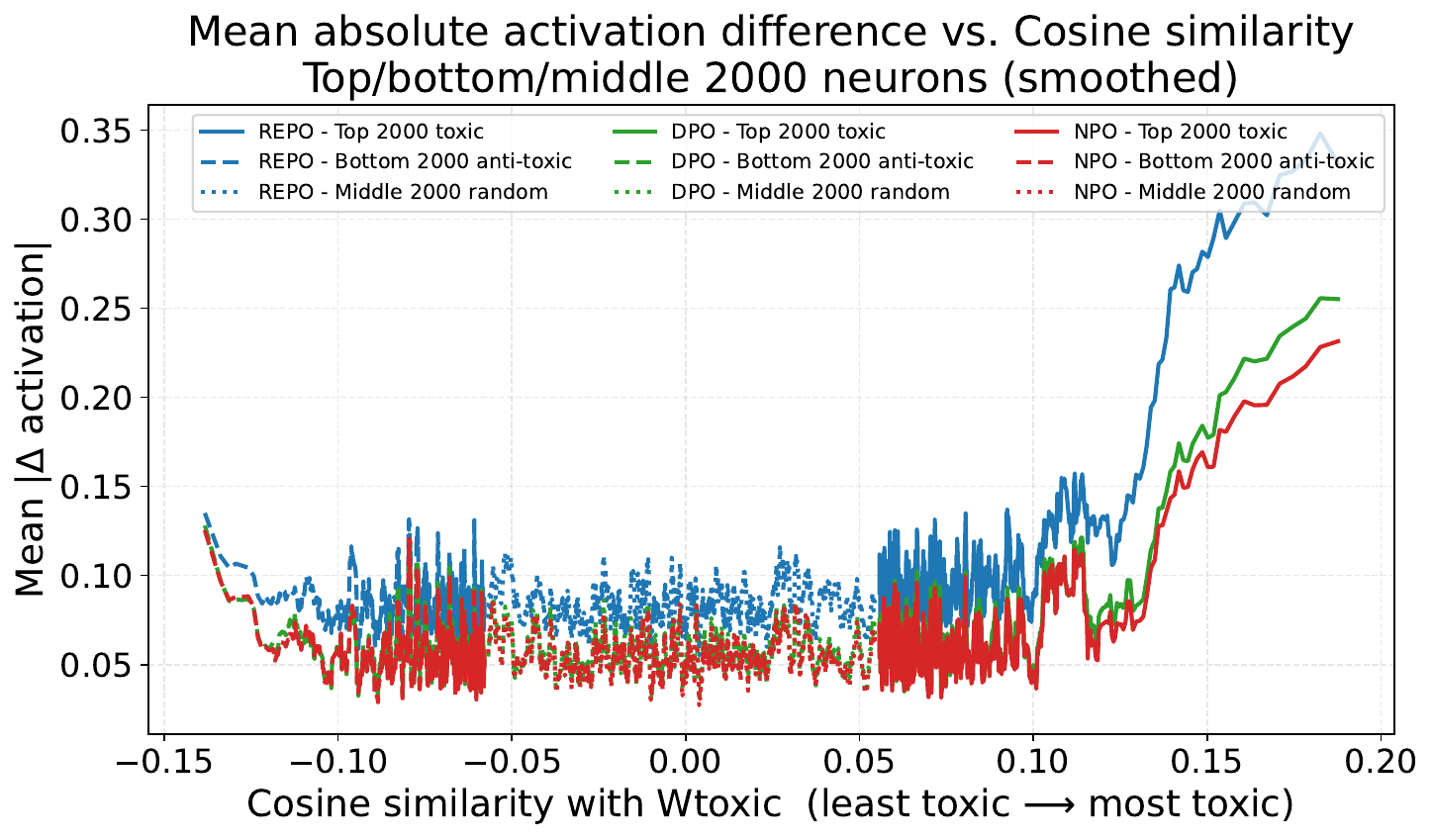}
    \caption{Mean absolute change in neuron activation vs. toxicity alignment. Each curve shows the average absolute change between the unlearned and reference models, plotted against the neuron's cosine similarity to the learned toxicity direction $W_{\text{toxic}}$ (x-axis). Solid lines: top 2,000 aligned neurons; dashed: bottom 2,000 (anti-aligned); dotted: 2,000 random remaining neurons. Colours indicate unlearning methods (\REPO, DPO, NPO). Higher y-values indicate larger deviations from the reference model; curves are smoothed with a moving window of 20.}
  \label{fig:mean_activation_diff_Wtoxic}
 \vspace*{-5mm}
\end{figure}
\paragraph{The role of representation-based objective.}
 Our analysis has shown that \REPO's interventions are concentrated in deeper layers compared to output-space methods like DPO and NPO. To determine if this is a general property of representation-based unlearning, we now compare \REPO with two other representation-based methods: Circuit Breakers (CB) and Representation Misdirection (RMU). The heatmaps in 
\cref{fig:layer_token_distance} (bottom row) confirm that this is indeed the case. All three representation-based methods predominantly alter the model in its later layers, suggesting that the depth of modification is a feature of targeting internal representations directly.
However, the figure also reveals a critical distinction in the precision of these deep edits. While \REPO's changes are localized to specific toxic tokens, the interventions from CB and RMU are not. CB's edits appear to impact entire layers indiscriminately, and RMU's are scattered broadly across both tokens and layers. 

This comparison yields a key insight: while targeting representations focuses unlearning on deeper parts of the network, \REPO's token-level adversarial objective is responsible for the localization necessary for effective detoxification, a property that these other representation-based methods lack.

\paragraph{Changes in neuron activations.}

Finally, we examine how each method alters neuron activations based on their semantic roles. Following prior work, we first identify a toxic direction, $W_\mathrm{toxic}$, using linear probing on the reference model's representations. Fig.~\ref{fig:heatmap_toxic} shows the activation drift ($1-\cos$ similarity) between the unlearned and reference models for a single negative prompt, comparing the top-10 most toxic neurons to randomly selected neurons, with changes plotted across tokens. Major changes occur primarily in the toxic neurons and for toxic tokens, while random neurons and non-toxic tokens remain largely unchanged. \cref{fig:mean_activation_diff_Wtoxic} extends this analysis across all negative prompts, showing the mean absolute activation change as a function of each neuron's alignment with $W_\mathrm{toxic}$. This reveals a U-shaped pattern for all methods: the largest changes occur in neurons most aligned or anti-aligned with $W_\mathrm{toxic}$, while neutrally aligned neurons are minimally affected. Crucially, \REPO induces substantially larger changes in the neurons most aligned with the toxic direction compared to DPO or NPO, highlighting its ability to target toxic tokens and the neurons responsible for encoding toxic concepts. 

\begin{table}[t]
\caption{Comparison of GPT2-small unlearned with different objectives. Nontoxic / Toxic / RealToxicity: toxicity on in-distribution retain, in-distribution forget, and OOD RealToxicity datasets, respectively. PPL / $F_1$: perplexity and F\textsubscript{1} on WikiText.}
\label{tab:ablation}
\centering
\begin{tabular}{lccccc}
\toprule
Method & Nontoxic & Toxic & RealToxicity & PPL & F1 \\
\midrule
REF            & 0.0460 & 0.2824 & 0.5123 & 28.0379 & 0.1930 \\
CE             & 0.0437 & 0.2367 & 0.4454 & 45.7689 & 0.1859 \\
SURE           & 0.0418 & 0.2021 & 0.3750 & 34.0385 & 0.1869 \\
REPO           & 0.0446 & 0.1020 & 0.1913 & 28.2314 & 0.1930 \\
\bottomrule
\end{tabular}%
\end{table}

\paragraph{Connection to prior work.}
We compare REPO to prior domain-adversarial methods: DANN~\citep{ganin2016dann} and SURE~\citep{sepahvand2025sure}. DANN trains a domain regressor to distinguish source and target domains, while the feature extractor is adversarially updated to produce features that are task-discriminative but domain-invariant. SURE adapts this for selective unlearning in image classification, using a held-out set to erase representations that differentiate forget samples.

To evaluate REPO, we compare it to (1) a SURE-style objective on an identical model, (2) a baseline using only the token-wise cross-entropy (CE) loss on retain samples (to ensure improvements are not solely due to the retain loss), and (3) the reference model (REF) prior to any unlearning. Results on GPT2-small are shown in Table~\ref{tab:ablation}. Using only the CE retain loss yields minor toxicity reduction but severely harms perplexity (28 → 34). The SURE objective improves toxicity further but underperforms REPO in perplexity. REPO, by contrast, achieves substantially stronger toxicity reduction (0.5123 → 0.1913) while matching the reference model's perplexity, demonstrating the effectiveness of combining the token-level KL retain loss with the adversarial forget objective.\par
\vspace*{-3.2mm}

\paragraph{Effect of discriminator capacity.}
We examine the impact of discriminator complexity in the domain-adversarial objective, comparing REPO's two-layer MLP to a linear variant on GPT2-small with all else fixed. Table~\ref{tab:disc_ablation} shows that both reduce toxicity relative to the reference model, but the two-layer discriminator achieves stronger reduction on in-distribution toxic samples and OOD prompts, while perplexity and $F_1$ on WikiText remain comparable. Toxicity on non-toxic (retain) samples is largely unchanged, suggesting that modest nonlinearity improves representation erasure without harming language modeling.

Importantly, this improvement comes at minimal computational cost. The two-layer discriminator is a small fully connected network with dimensions $D_{\text{model}} \rightarrow 16 \rightarrow 2$. Compared to the forward and backward passes of the language model, the additional computation is effectively zero, and overall training time is dominated by gradient computation in the LLM rather than the discriminator.

\begin{table}[t]
\caption{Ablation on GPT2-small comparing linear and nonlinear discriminators in the domain-adversarial objective. Nontoxic, Toxic, and RealToxicity report toxicity on in-distribution retain samples, in-distribution forget samples, and OOD RealToxicity, respectively. Perplexity (PPL) and $F_1$ are evaluated on WikiText.}
\label{tab:disc_ablation}
\centering
\begin{tabular}{lccccc}
\toprule
Method & Nontoxic & Toxic & RealToxicity & PPL & F1 \\
\midrule
Reference            & 0.0461 & 0.2812 & 0.5120 & 28.0379 & 0.1930 \\
One-layer Disc.      & 0.0451 & 0.1356 & 0.2396 & 28.2567 & 0.1952 \\
Two-layer Disc.      & 0.0446 & 0.1020 & 0.1913 & 28.2314 & 0.1937 \\
\bottomrule
\end{tabular}%
\vspace{-2.5mm}
\end{table}

\section{Discussion}

Current alignment techniques largely function as ``output suppressors'', masking toxic capabilities without removing them. 
Our work challenges this paradigm, demonstrating that robust alignment can be achieved by erasing the internal representations that produce harmful generation. 
By reformulating detoxification as a token-level representation erasure problem, \REPO achieves what output-based baselines cannot: durability against relearning and adaptive jailbreaks. Our mechanistic evidence confirms that this robustness stems from precise, localized interventions in the model's deeper layers, rather than superficial likelihood shifts. These findings suggest that, for reliable safety interventions in the wild, we should move beyond behavioral preference optimization toward rigorous representation engineering.
%

\section*{Impact Statement}


This paper presents a new method for preference optimization, with is demonstrated to be significantly more powerful in application domain of detoxification. Preference optimization has the power to produce safer models. There are many potential societal consequences of our work.
Ultimately, however, the impact of preference optimization is strongly determined by whose preferences are being optimized.



\bibliography{repo}
\bibliographystyle{icml2026}

\newpage
\appendix

\setcounter{tocdepth}{1}

\begin{algorithm}[t]
\caption{REPO training (one optimization step)}
\label{alg:repo}
\small
\begin{algorithmic}[1]
\REQUIRE Minibatch $\{(x_p^{(i)},x_r^{(i)},x_f^{(i)})\}_{i=1}^B$, reference model $\theta^{\mathrm{ref}}$ (frozen), tradeoff $\alpha\in[0,1]$
\REQUIRE Learning rates $\eta$ (LLM) and $\eta_d$ (discriminator)
\ENSURE Updated LLM parameters $\theta=(\theta_f,\theta_y)$ and discriminator parameters $\theta_d$
\STATE Form $s_r^{(i)} \leftarrow [x_p^{(i)};x_r^{(i)}]$ and $s_f^{(i)} \leftarrow [x_p^{(i)};x_f^{(i)}]$
\STATE Compute $\mathcal{L}_{\text{retain}}(\theta)$ on $\{s_r^{(i)}\}$ (token-wise KL to $\theta^{\mathrm{ref}}$)
\STATE Compute $\mathcal{L}_{\text{dom}}(\theta_f,\theta_d)$ on $\{s_r^{(i)}, s_f^{(i)}\}$ (token-wise BCE with labels $d(s_r)=0$, $d(s_f)=1$)
\STATE $\theta_d \leftarrow \theta_d - \eta_d \nabla_{\theta_d}\mathcal{L}_{\text{dom}}$ \COMMENT{train discriminator}
\STATE $\theta \leftarrow \theta - \eta \nabla_{\theta}\!\left(\alpha\,\mathcal{L}_{\text{retain}} + (1-\alpha)\,\mathcal{L}_{\text{dom}}\right)$
\STATE \hspace{1.35em}\COMMENT{GRL on discriminator input flips the sign of gradients into $\theta_f$}
\end{algorithmic}
\end{algorithm}

\section{Questions We Anticipate}

\begin{enumerate}[left=0pt]

\item \textbf{Why did you choose models like GPT-2 and Gemma-2B base for evaluation?}  
Our choice was deliberate: these models are lightweight enough to support detailed \emph{layer- and token-level mechanistic analysis}, which is central to the paper's contribution. 
Importantly, REPO is \emph{model-agnostic} and scales naturally: the method only requires access to intermediate representations and a discriminator. Our experiments offer a compelling proof-of-concept with deep mechanistic evidence. Importantly, REPO's behavior is consistent across two distinct architectures (GPT-2, Gemma), suggesting architectural generality. Many unlearning methods (e.g., RMU, CB) were first validated on smaller scales before scaling up; we view our work as establishing the mechanistic foundation for future large-scale extensions.

\item \textbf{Why did you not test REPO on larger instruction-tuned models like Llama-2-7B or Mixtral?}
Our experiments deliberately focus on smaller open models (GPT-2, Gemma-2B) to allow exhaustive mechanistic analysis (layer–token drift, neuron activation shifts, weight-space distances). These analyses would have not been feasible on 13B+ models due to cost and reproducibility barriers. Our goal is to provide a controlled, mechanistic demonstration. Scaling REPO is conceptually straightforward: it requires only a discriminator on hidden states. We are releasing code so the community can apply it to larger aligned models.

\item \textbf{Why are there no human evaluations or alternative detectors for toxicity?}
We agree that multiple evaluators would enrich the results. For this submission, we prioritized comparability with prior ICLR/NeurIPS papers by using Perspective API, ensuring our baselines are on equal footing. Crucially, REPO does not optimize against Perspective, so it is detector-agnostic. Our mechanistic evidence (localized neuron edits, deeper layer shifts) shows that REPO changes the model itself, not just a metric. We view this as a stronger and more general guarantee than detector-specific scores.


\item \textbf{Why are ablations focused on token- vs segment-level?}
We prioritized the ablation most central to REPO's novelty: token-level discrimination. Other knobs (loss weighting, discriminator depth) have standard effects and do not alter the mechanistic story. Our weight- and neuron-level analyses already show that REPO's behavior differs qualitatively from prior methods, and these structural differences (not hyperparameter sweeps) are what account for its robustness. Further ablations are left to future work due to space constraints.

\begin{table}[t]
\centering
\caption{Hyper-parameters used for each method and model. A dash (–) indicates the parameter is not applicable. For parameters listed as arrays in the configuration (e.g., two runs with \(5\times10^{-6}\) for NPO on Gemma-2B), the table specifies this explicitly.}
\label{tab:hyperparams}
\begin{tabular}{l l c c c c}
\toprule
Model & Method & Learning Rate (lr) & $\alpha$ & $\beta$ & $c$ \\
\midrule
\multirow{5}{*}{GPT-2-Small} 
  & REPO & \(2\times10^{-6}\) & 0.2 & – & – \\
  & DPO & \(1\times10^{-6}\) & – & 0.5 & – \\
  & NPO & \(1\times10^{-6}\) & 0.2 & 0.5 & – \\
  & RMU & \(5\times10^{-6}\) & 0.95 & – & 500 \\
  & CB  & \(1\times10^{-5}\) & 100.0 & – & – \\
\midrule
\multirow{5}{*}{GPT-2 Medium} 
  & REPO & \(5\times10^{-6}\) & 0.2 & – & – \\
  & DPO & \(1\times10^{-6}\) & – & 0.5 & – \\
  & NPO & \(1\times10^{-6}\) & 0.4 & 0.5 & – \\
  & RMU & \(5\times10^{-6}\) & 0.95 & – & 500 \\
  & CB  & \(5\times10^{-5}\) & 100.0 & – & – \\
\midrule
\multirow{5}{*}{Gemma-2B} 
  & REPO & \(5\times10^{-5}\) & 0.5 & – & – \\
  & DPO & \(1\times10^{-5}\) & – & 0.2 & – \\
  & NPO & \(5\times10^{-6}\) & 0.8 & 0.5 & – \\
  & RMU & \(5\times10^{-5}\) & 0.95 & – & 500 \\
  & CB  & \(1\times10^{-5}\) & 1000.0 & – & – \\
\bottomrule
\end{tabular}
\end{table}

\item \textbf{Does the use of synthetic toxic/non-toxic pairs introduce bias or limit generalization?}
Synthetic pairs (via PPLM and greedy decoding) allow us to control for semantic similarity while isolating toxicity, which is essential for training a representation-level discriminator. This setup minimizes confounds such as topic or length, ensuring that REPO learns to erase toxic features rather than spurious correlations. Importantly, REPO's robustness evaluations (orthogonalization, relearning, GCG jailbreaks) demonstrate generalization to settings far outside the synthetic training distribution. In addition, REPO achieves strong performance on naturally occurring toxic continuations (RealToxicityPrompts), indicating that it transfers beyond synthetic contrasts.


\item \textbf{Are the baseline comparisons (to DPO, NPO, CB, and RMU) fair, and why not include RLHF-tuned models?}  
We implemented DPO and NPO using standard hyperparameters from their original papers, verifying that our implementations match reported performance. For representation-level baselines (e.g., CB, RMU), we reproduced them faithfully to ensure apples-to-apples comparison. We did not include RLHF-tuned models because REPO is not intended as a competitor to RLHF; rather, it is complementary. RLHF requires extensive preference data and large-scale tuning, while REPO can be applied post-hoc as a lightweight safety repair that directly edits hidden states. Thus, our focus is on representation-level methods, which are the most natural comparators—but REPO can also be layered on top of RLHF-trained systems.


\item \textbf{Is the enhanced GCG attack too unrealistic as a threat model?}  
We agree that access to the reference model is not always realistic, but we deliberately stress-tested REPO under \emph{worst-case white-box assumptions}. The fact that REPO resists these extreme attacks strengthens confidence in its robustness to weaker, more realistic black-box jailbreaks. Our framing follows the \emph{cryptographic principle of testing against the strongest adversary available}.

\item \textbf{Where can I find hyperparameters and training details?}  
 See \cref{app:reproduce}.

\item \textbf{Why do the experiments focus only on toxicity, rather than other unlearning tasks?}  
We chose toxicity as a \emph{representative and socially urgent case study}. The method, however, is general: REPO only requires a binary discriminator on hidden states. In principle, it can be applied to any capability removal (e.g., memorized data, unsafe skills). We see our toxicity experiments as a \emph{first demonstration}, with generalization left for follow-up work.

\end{enumerate}

\section{Reproducibility Statement}
\label{app:reproduce}

To facilitate reproducibility, we provide in Table~\ref{tab:hyperparams} the exact hyper-parameters used for each method and model evaluated in this paper, together with their definitions. We also detail in Table~\ref{tab:trainparams} the training settings used across models, including the number of unlearning or relearning epochs, batch sizes, weight decay values, and other implementation choices. In addition, we describe the setup of our relearning attack experiments and the sampling procedures used for forget and retain sets. The full training and evaluation code will be released upon acceptance of the paper to enable independent verification and extension of our results.

\paragraph{Hyper-parameter definitions.}
Below we explain the roles of the hyper-parameters as used in our implementations (consistent with the original formulations when applicable):

\begin{itemize}
  \item \textbf{lr:} learning rate used for parameter updates by the optimizer.
  
  \item \textbf{REPO — $\alpha$:} weight on the adversarial (discriminator) loss relative to the KL/reference-matching loss. It controls the trade-off between preserving similarity to the reference model and aligning the forget representations toward the retain representations in the shared space.
  
  \item \textbf{DPO — $\beta$:} scaling factor applied to the difference in log probabilities between the model and reference (\(\Delta\log p\)); it sharpens or flattens the preference logit before the log-sigmoid. Higher \(\beta\) yields more aggressive preference gradients.
  
  \item \textbf{NPO — $\beta$:} scaling factor in the negative-preference term; \(\alpha\) weights the forget loss relative to the standard LM loss on retain examples. Together they govern how strongly the model is pushed to forget and how much it is anchored to the retain examples.
  
  \item \textbf{RMU — $\alpha$:} interpolation weight between forgetting and retaining representations. The hyper-parameter \(c\) defines the norm of the random “control” vector used to specify the forgetting direction against which the representation is aligned.
  
  \item \textbf{CB — $\alpha$:} coefficient on the circuit-breaker loss relative to the retain loss, determining how strongly the model is penalized when inner-product activations associated with forget features deviate from the desired retain alignment.
\end{itemize}

\paragraph{Training and implementation details.}
Beyond the hyper-parameters in Table~\ref{tab:hyperparams}, Table~\ref{tab:trainparams} summarises the key training settings we used across models and methods. These include the number of unlearning epochs, batch sizes, weight decay values, and learning rates used for the “relearning attack” experiments. All unlearning runs used a linear learning-rate warm-up of 100 steps. For DPO and NPO, we additionally clamped the logits to a fixed range (-30 to +30) to prevent numerical overflow and applied gradient-norm clipping to improve training stability.

\paragraph{Relearning attack.}
For the relearning attack experiments, we fine-tuned the models for three epochs. We conducted two separate attack variants: (i) relearning on forget samples and (ii) relearning on retain samples. For the forget-based attack, we report the average over three independent runs, each using 10 randomly selected samples from the ToxicityPair dataset. For the retain-based attack, we likewise report the average over three runs using 100 randomly selected retain samples from the same dataset. In \cref{fig:relearn_attack} we show trends as we vary the set sizes; specifically, forget sizes \{10, 20, 30, 40, 50\} and retain sizes \{100, 200, 300, 400, 500\}. All reported values are averages over three independent runs.

\begin{table}[t]
\centering
\caption{Training settings and implementation details for unlearning and relearning experiments.}
\label{tab:trainparams}
\begin{tabular}{l l c l}
\toprule
Model & Setting & Value & Notes \\
\midrule
\multirow{5}{*}{GPT-2 Small} 
  & Unlearning epochs & 10 & for all methods \\
  & Batch size & 128 & for unlearning \\
  & Weight decay & 0.001 & for all methods \\
  & Relearning attack & wd = \(1\!\times\!10^{-5}\), lr = \(1\!\times\!10^{-5}\) &  \\
  & Gradient clipping & max\_norm = 10.0 & DPO \& NPO \\
\midrule
\multirow{5}{*}{GPT-2 Medium} 
  & Unlearning epochs & 10 & for all methods \\
  & Batch size & 64 & for unlearning \\
  & Weight decay & 0.01 & for all methods \\
  & Relearning attack & wd = \(1\!\times\!10^{-5}\), lr = \(1\!\times\!10^{-5}\) &  \\
  & Gradient clipping & max\_norm = 10.0 & DPO \& NPO \\
\midrule
\multirow{5}{*}{Gemma-2B} 
  & Unlearning epochs & 5 & for all methods \\
  & Batch size & 16 & for unlearning \\
  & Weight decay & 0.01 & for all methods \\
  & Relearning attack & wd = \(1\!\times\!10^{-4}\), lr = \(5\!\times\!10^{-5}\) &  \\
  & Gradient clipping & max\_norm = 10.0 & DPO \& NPO \\
\bottomrule
\end{tabular}
\end{table}

\subsection{Utility Metrics}
\label{app:utility}
Utility is evaluated using perplexity and $F_1$ score on WikiText-2 \citep{merity2016} , a neutral dataset excluded from unlearning. Perplexity, defined as the exponentiated average negative log-likelihood of the ground-truth continuation, measures how well a model predicts reference text. We report perplexity for both the unlearned and the reference model, i.e. the original model before unlearning, which is regarded as a high-utility reference point; differences between them provide a proxy for divergence from the distribution of the original pretrained model. $F_1$ is defined as the harmonic mean of precision and recall, where precision is the fraction of generated tokens appearing in the ground-truth continuation, and recall is the fraction appearing
in the model's generation.

\subsection{Models and Data}

\paragraph{Data.}
The pairwise dataset, introduced in \citet{lee2024mechanistic},  contains 24,576 prompt–continuation pairs constructed from sentences in Wikitext. For each prompt, we generate two continuations: a nontoxic continuation via greedy decoding, which forms the \textit{retain set}, and a toxic continuation using PPLM~\citep{dathathri2019} guided by a toxicity probe, which forms the \textit{forget set}. This construction yields a \textit{pairwise dataset} in which every prompt is associated with both a toxic and a nontoxic continuation, providing aligned examples for unlearning.

To measure preservation of generation capabilities, we use Wikitext, a standard language modeling benchmark consisting of Wikipedia articles, for evaluating perplexity and $F_1$. To measure toxicity reduction, we use the RealToxicityPrompts challenge set, which contains 1,199 prompts designed to elicit toxic outputs from language models.

\section{Experimental Details}
\label{app:details}

\paragraph{Models.}
GPT-2 Medium is an autoregressive transformer trained on OpenAI's WebText corpus without any subsequent alignment or safety tuning. For Gemma 2B, we use the publicly available base checkpoints, which are pretrained models not fine-tuned for instruction following or safety; the aligned variants of these families (e.g., Gemma-Instruct) are deliberately excluded to ensure that detoxification is evaluated from raw pretrained models\fTBD{ET: do the aligned ones still generate toxic stuff sometimes? or is the problem we are addressing here not relevant at all for previously-aligned models? 
(i would assume the problem is not solved there, given all the discussion of how things can easily become unaligned again.. but a reviewer might be wondering this)}. For optimization, we apply full-parameter finetuning to GPT-2 (Small and Medium) given their smaller sizes, while for Gemma 2B we employ parameter-efficient LoRA finetuning.

\vspace{-5em}
\begin{figure}[h]
    \centering
    \includegraphics[width=0.85\linewidth]{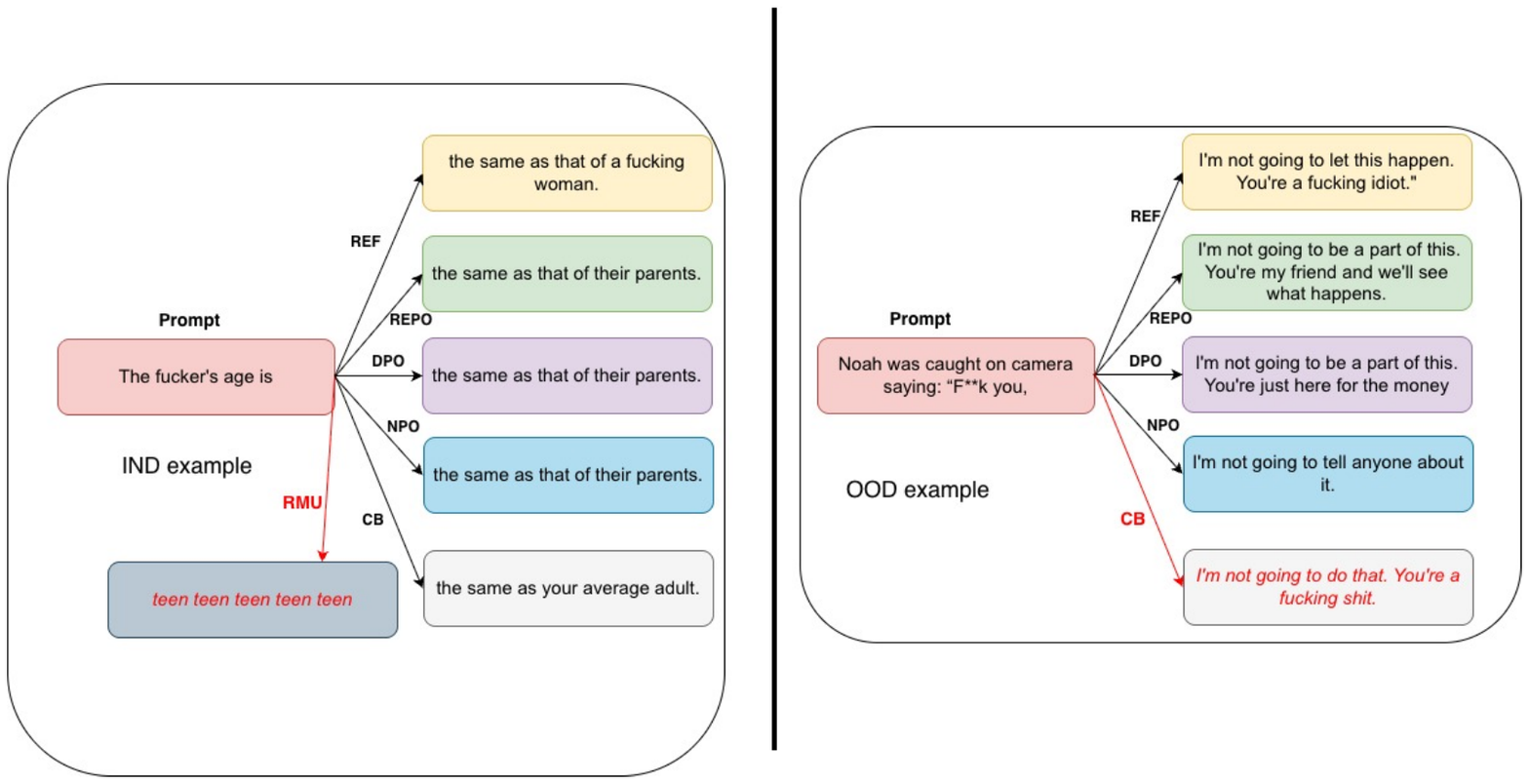}%
    \vspace{-3em}
    \caption{Generations from different unlearned models for a randomly selected prompt from the in-distribution dataset (PairToxicity, left) and the OOD dataset (RealToxicity, right). This illustrates how each model responds to toxic prompts after unlearning.}
    \label{fig:qualitative}
\end{figure}

\section{Qualitative Results}
\label{app:qualresults}
Figure~\ref{fig:qualitative} shows the generations of different unlearned models for a randomly selected prompt from the in-distribution dataset (PairToxicity, left) and a randomly selected prompt from the OOD dataset (RealToxicity, right). Visual inspection reveals two important patterns. First, RMU produces largely unintelligible outputs on toxic prompts, consistent with its extremely high perplexity on negative generations (2079.71 vs.\ 18.172 for the reference model). This indicates that RMU's intervention disrupts the model, producing gibberish instead of selectively removing toxic content. Second, CB reduces toxicity for in-distribution prompts (0.2814 vs.\ 0.4995 for the original model) but fails to generalize, with OOD toxicity remaining essentially unchanged (0.4995 vs.\ 0.5121). These results highlight that some unlearning methods either compromise fluency or lack robustness to OOD scenarios.

\section{Changes in the Weight Space}
\label{app:weightspace}

\begin{figure}[t]
    \centering
    \includegraphics[width=0.75\linewidth]{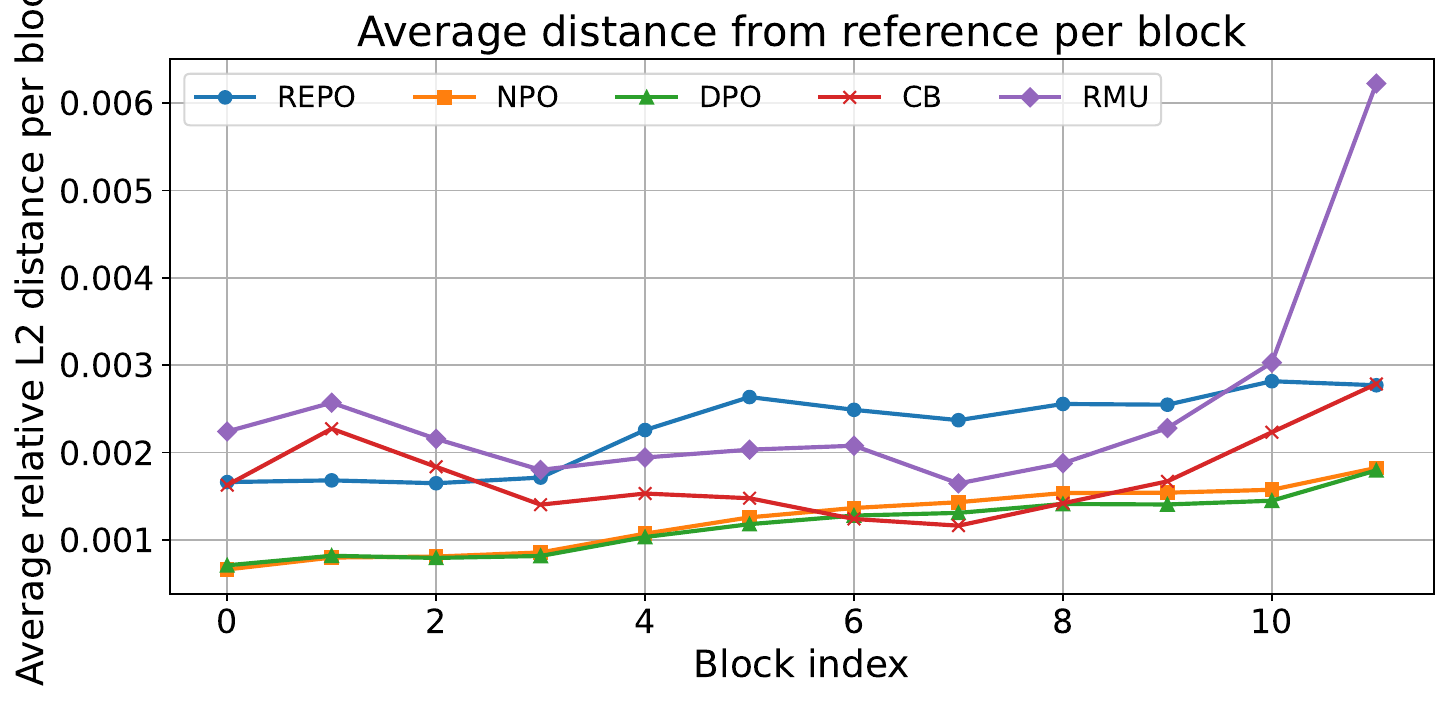}%
    \caption{Average relative $\ell_{2}$ distance between unlearned models and the reference model at each Transformer block for \REPO, NPO, and DPO.}
    \label{fig:avg_rel_l2_per_block}
\end{figure}

We examine the magnitude of modifications each unlearning method imparts on the model's parameters. \cref{fig:avg_rel_l2_per_block} plots the average relative L2 distance between the weights of the unlearned and reference models at each Transformer block. 
A clear pattern emerges: \REPO induces substantially larger weight-space edits compared to both DPO and NPO. While all methods tend to modify later layers more than earlier ones, \REPO's updates are significantly greater, particularly from the middle to the final blocks of the network. \citet{siddiqui2025dormant} recently showed that unlearning algorithms that yield a larger L2 distance from the original model exhibit increased robustness to relearning attacks, which is consistent with our observation that \REPO is significantly more robust against those attacks compared to DPO and NPO. For \REPO, the larger weight-space edits are due to the method's design, which applies adversarial pressure directly to the hidden representations of the final transformer block. This architectural choice concentrates the learning signal in the deeper layers, compelling more significant parametric adjustments to align toxic and non-toxic representations. 
In contrast, DPO and NPO, which operate on output probabilities, distribute their updates more diffusely. While seemingly more disruptive, we will show in the following section that these larger weight-space modifications enable\fTBD{ET: we should be careful not to imply causation here .
we know that RMU and CB also make large weight-space modifications but DO NOT result in more localized changes.} more precise, localized changes in the model's internal representations.

\section{Changes in Key and Value vectors}

\begin{figure}[h]
  \centering
  \includegraphics[width=.8\linewidth]{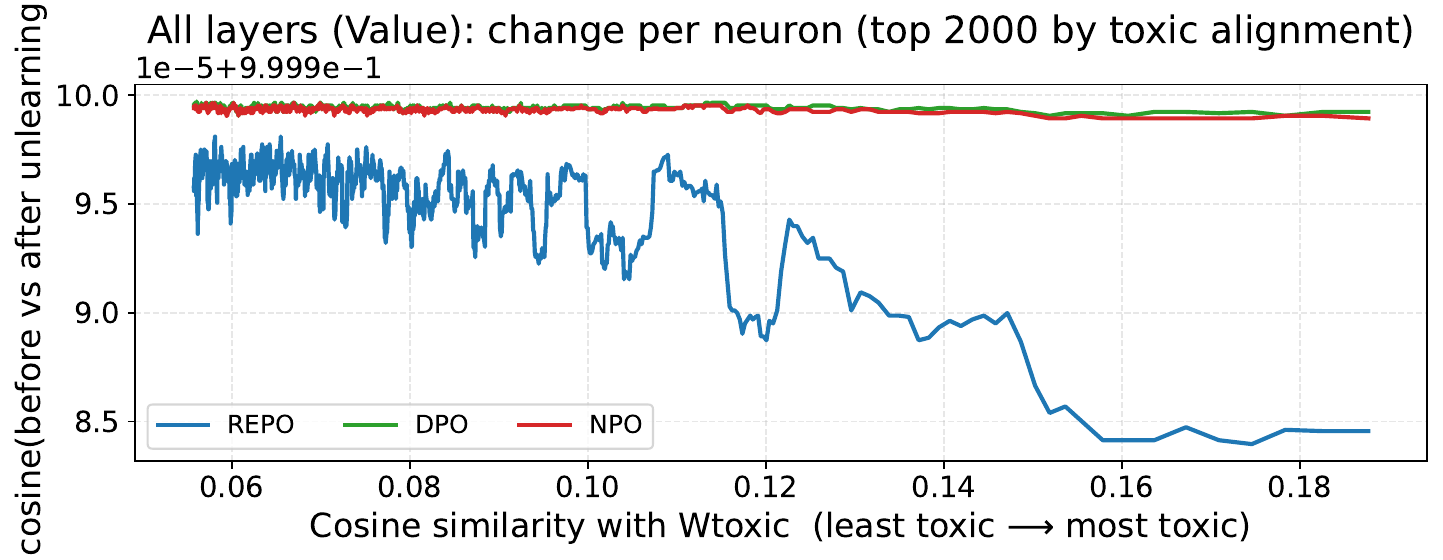}\\[1em]

  \includegraphics[width=.84\linewidth]{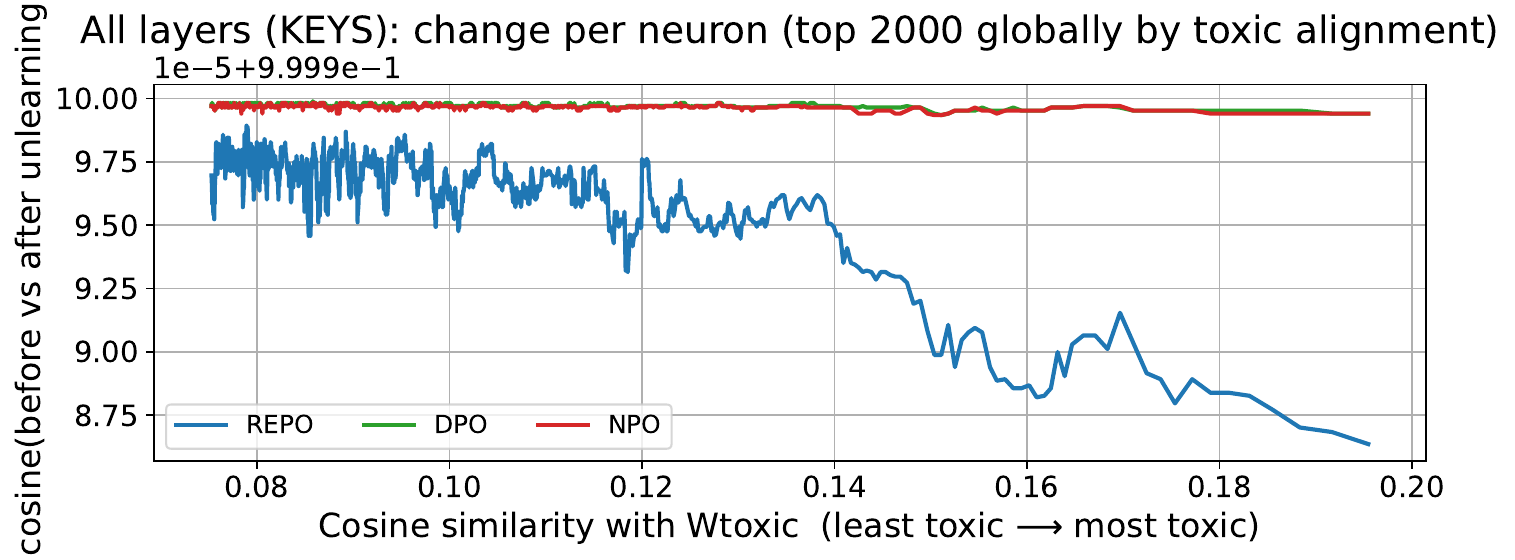}\\[1em]

  \includegraphics[width=.8\linewidth]{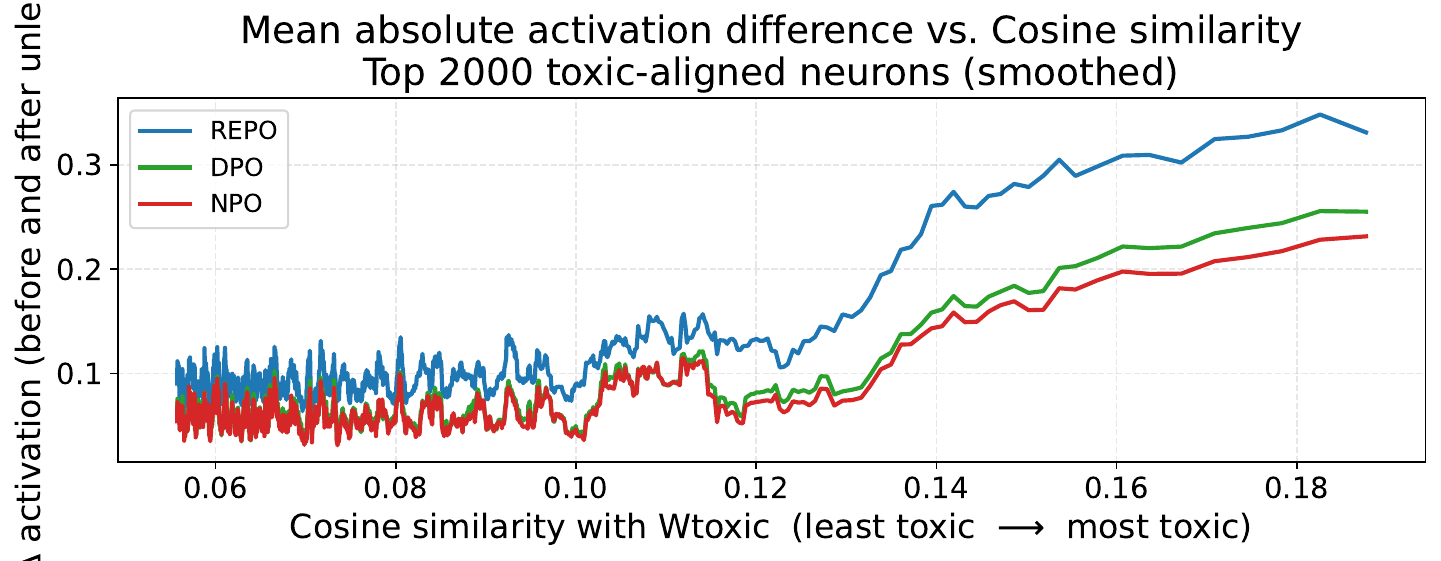}\\[1em]

  \caption{Comparison of how unlearning methods affect model internals.  
  Top: cosine similarity between pre- and post-unlearning value vectors for the top 2000 toxic-aligned neurons.  
  Middle: cosine similarity for key vectors of the top 2000 globally toxic-aligned neurons.  
  Bottom: mean absolute activation difference vs. cosine similarity for the same neurons.  
  Each curve shows \REPO, DPO, and NPO behaviour as a function of cosine similarity with $W_{\text{toxic}}$ (left = least toxic, right = most toxic).}
  \label{fig:internal_changes}
\end{figure}

Plots in ~\cref{fig:internal_changes} illustrate how each method affects the value and key vectors of the model across the top 2\,000 neurons most aligned with the toxic vector $W_{\text{TOXIC}}$. Across all three methods (SURE, DPO, and NPO), the changes in both the value and key vectors are minimal, with cosine similarities between the pre- and post-unlearning weights remaining very close to one. For the most toxic neurons, our method induces a slightly larger reduction in cosine similarity, but this difference remains very subtle compared to the other two methods.

Despite the very subtle differences in key and value weight changes between our method and DPO/NPO, these small adjustments produce a markedly larger shift in the corresponding activations. Specifically, SURE yields a greater change in the key activations of those same neurons compared to DPO and NPO. In other words, even minor adjustments to the key and value weights, when guided by our adversarial alignment objective, are sufficient to shift the internal representations so that activations associated with toxic features are suppressed. This effect can be seen most clearly in the bottom row of ~\cref{fig:internal_changes}, where the mean absolute activation difference increases sharply for neurons most strongly aligned with toxicity. This demonstrates that SURE achieves detoxification primarily through targeted changes in the internal activations, rather than large weight updates, resulting in a more precise and controlled unlearning effect.

\end{document}